# Path Planning Algorithms for Robotic Aquaculture Monitoring

Anthony Davis*, Srijita Mukherjee, Paul S. Wills, Bing Ouyang
Harbor Branch Oceanographic Institute
Florida Atlantic University, Fort Pierce, Florida, USA

## ABSTRACT

Aerial drones have great potential to monitor large areas quickly and efficiently. Aquaculture is an industry that requires continuous water quality data to successfully grow and harvest fish. The Hybrid Aerial Underwater Robotic System (HAUCS) is designed to collect water quality data of aquaculture ponds to reduce labor costs for farmers. The routing of drones to cover each fish pond on an aquaculture farm can be reduced to the Vehicle Routing Problem. A dataset is created to simulate the distribution of ponds on a farm and is used to assess the HAUCS Path Planning Algorithm (HPP). Its performance is compared with the Google Linear Optimization Package (GLOP) and a Graph Attention Model (AM) for routing problems. GLOP is the most efficient solver for 50 to 200 ponds at the expense of long run times, while HPP outperforms the other methods in solution quality and run time for instances larger than 200 ponds.

## 1. INTRODUCTION

Aquaculture as an industry has experienced exponential growth in the past 5 decades [1]. To meet rising consumer demand for fish and maintain affordable prices, the aquaculture industry must improve its labor efficiency through automation. One of the most critical aspects of the pond aquaculture is to measure proper dissolved oxygen (DO) level in each of the pond. Traditionally, DO monitoring relies on human operators driving trucks through the farm to manually measure the DO concentration in each pond, which is both labor-intensive and costly. While the state-of-the-art pond-buoy based sensor installation helps to alleviate the burden on the human operators, such solutions may suffer high cost, poor scalability and maintenance difficulty due to biofouling. The Hybrid Aerial Underwater Robotic System (HAUCS) is a system designed to periodically monitor DO levels in ponds, a normally labor-intensive process [2]. In the HAUCS framework, to mitigate the challenges in the aforementioned monitoring solutions, a team of drones with underwater DO sensors are dispatched to the aquaculture ponds to measure their DO levels at a regular time interval. This design will reduce maintenance costs at the expense of added design complexity necessary to coordinate the drones.

This team of drones must be given appropriate routes to maintain efficiency and low cost. By representing pond locations as nodes on a graph, this can be described as the Vehicle Routing Problem (VRP), a well-studied generalization of the Traveling Salesman Problem (TSP). Unlike TSP, VRP algorithms have the added complexity of finding the most optimal collection of routes such that the longest route is minimized, as opposed to simply finding the shortest route to visit the required locations.

Current VRP techniques typically focus on instances with less than 100 locations, and there are relatively few methods for solving VRP with a large number of locations. However, commercial aquaculture farms may manage 1,000 or more ponds. Additionally, most VRP studies use randomly generated node locations, but aquaculture pond locations are not placed randomly. They are placed in a dense grid according to available land. These unique circumstances are replicated in our data by laying out a grid of points within a randomly generated polygon and removing 20% of the points for variability.

The HAUCS Path Planning (HPP) algorithm intends to take advantage of this regular structure to quickly and effectively plan routes for a team of drones on an aquaculture farm. With the assumption that nodes are generally placed on a grid and given a large enough graph, heuristics can be constructed that are efficient, quick to execute, and simply understood by human operators. We compare our work with other approaches to establish how HPP compares on three important metrics: total distance, maximum route distance, and run time.

*anthonydavis2020@fau.edu

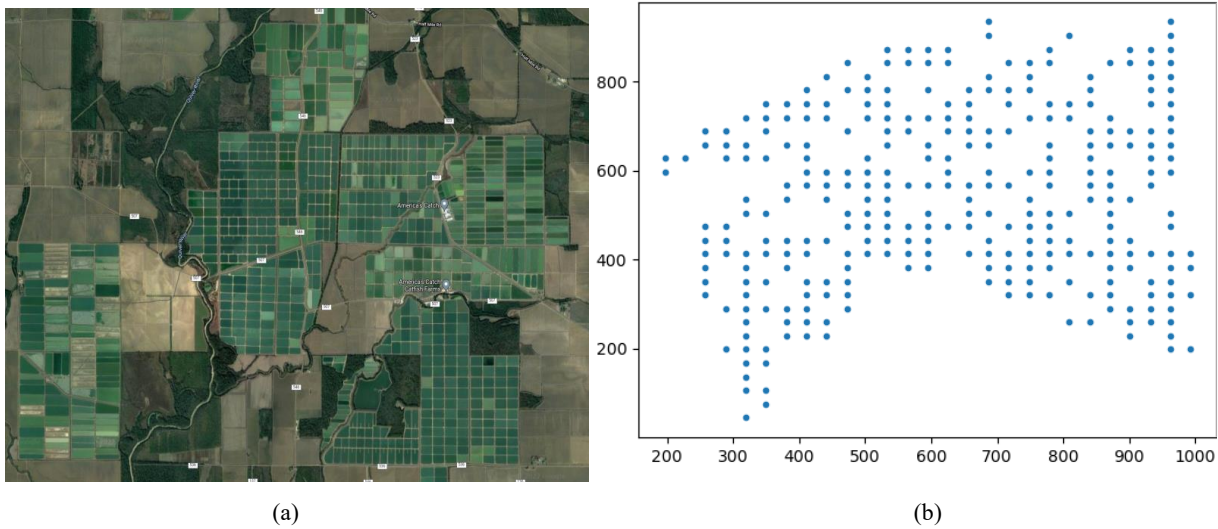

(a)                       (b)

Figure 1: (a) America's Catch Catfish Farm with approximately 700 ponds. (b) Example 300 node instance showing simulated aquaculture pond distribution

## 2. PREVIOUS WORK

TSP is perhaps the most well-studied combinatorial optimization problem in computer science. However, VRP is more realistic for practical applications and has attracted significant research. There are numerous variations of VRP, such as the Capacitated VRP where vehicles have a limited capacity and customers have a certain demand, VRP with Time Windows, where locations must be visited within specific times, and Distance Constrained VRP, where vehicles have a limited range before returning to the depot. VRP with no additional constraints is sometimes referred to as the Multiple Traveling Salesman Problem (MTSP). Several surveys specifically addressing MTSP and VRP go into further detail about the current literature's approaches and applications [3][4][5].

Modern approaches to combinatorial optimization problems such as VRP span several computer science disciplines, most notably linear optimization, metaheuristics, and machine learning. Other approaches such as exact algorithms are well studied but have long run times [6], and the study of simple heuristics has stalled mainly since the development of more sophisticated algorithms.

### 2.1. Linear Optimization and Exact Solutions

Linear optimization tends to find optimal solutions but has at least polynomial computational complexity and therefore cannot be used on large graphs [7]. This is an attractive method when working with small enough graphs, but otherwise may be intractable. Operations research efforts have led to the release of open-source tools for VRP, such as Google's OR-Tools [8] which enable easy implementation of VRP solvers.

Exact solutions are less practical for our purposes because of their high computational complexity, but Almoustafa et al. proposed an exact method for distance-constrained VRP solutions on large graphs [9], which is highly applicable to the HAUCS use case. Others such as Pecin, Qureshi, Queiroga, and many more propose alternative exact solutions for VRP [10][11][12].

## 2.2. Metaheuristics

Metaheuristics are highly successful techniques that are widely applicable because of their general lack of assumptions, but can be unreliable in that they may not have stable convergence and may have unknown run times [13]. These are techniques such as ant colony optimization [14], genetic algorithms [15], simulated annealing [16], and adaptive large neighborhood search [17]. There is a great deal of literature on effective metaheuristics for VRP and other optimization problems [18]. Vidal et al. propose a unified framework for VRP metaheuristics which is highly optimal and efficient [19].

## 2.3. Machine Learning

The vast number of VRP variations makes it challenging to find rule-based methods that apply to a specific problem, but some work has been done to produce generalizable methods using machine learning algorithms. Machine learning has proven to be highly effective for a wide range of applications but suffers from specialized hardware requirements, lack of explainability, and many of the same problems as metaheuristics, including unreliable convergence and unpredictable runtimes.

Graph neural networks are developing to become extremely useful in many domains of science, particularly chemistry and biology, and have shown use as a means to coordinate multi-agent systems [20] as well as path planning [21]. Kool et al. use a graph attention network to encode node locations and a reinforcement learning algorithm to train the navigation network, resulting in a method that can solve multiple types of routing problems [22]. Other machine learning approaches include Neural Large Neighborhood Search [23], Residual Graph Attention Networks [24], Reinforcement Learning for TSP [25], and more.

## 2.4. Drone specific approaches

Additionally, much research has been done on VRP pertaining specifically to aerial drones. The study of using drones to navigate a TSP scenario is motivated primarily for commercial home deliveries [26]. Rigas et al. use linear optimization to solve scheduling problems with a fleet of drones on a monitoring mission [27]. Raj and Kim et al. consider the flying sidekick problem, allowing drones to land on a truck to save battery and increase range [28][29]. Kitjacharoenchai et al. use heuristics to model MTSP for drones [30]. Es Yurek and Ozmutlu investigate TSP with a recharging policy for drones [31]. Drone speed is considered by Tamke and Buscher as an additional VRP constraint [32]. Cavani et al. propose multiple exact methods for MTSP, specifically for drones [33].

## 3. METHODS

HPP is an area coverage path planning strategy that enables the HAUCS platforms to periodically monitor aquaculture farms in a back-and-forth pattern. It can be described as a two-phase simple heuristic method. It first clusters, then routes. Clustering is accomplished using k-means on the entire node set to produce evenly spaced clusters. The convex polygons surrounding each cluster are then used in the routing phase to determine the path for an individual drone. The underlying assumption that must be met for this clustering to work is that the distribution of nodes is generally even. If nodes are placed randomly and have areas of higher density, this assumption would not hold and HPP would likely fail to produce efficient routes.

The antipodal pairs of each convex polygon surrounding the clusters are calculated using the Shamos algorithm [34]. Antipodal points are vertices of a convex polygon that admit parallel lines of support such that the whole polygon lies in-between. This set of antipodal pairs from each pair of vertices is then fed to the Optimal Coverage Algorithm (OCA) by Mukherjee et al. [35]. OCA calculates the shortest optimal back and forth path that starts and ends from one of the antipodal point pairs. The route is then assigned to follow the optimal back-and-forth pattern as closely as possible.

If a cluster happens to be too small to form a polygon, for example, if a cluster is located on a peninsula and only contains 1 or 2 points, then the subsequent closest nodes are taken into that cluster from its neighboring cluster until all clusters have a valid size and shape. Figure 2 demonstrates the evenly spaced clusters of approximately equal size despite the irregular outline shape of the node locations.

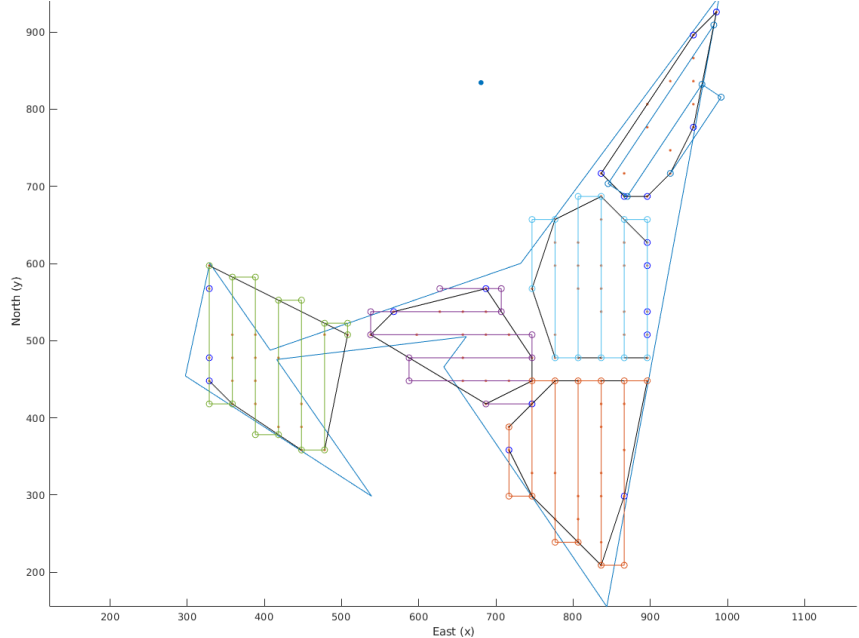

Figure 2: Plot of HPP clusters and optimal back and forth route plans. The blue dot signifies the depot and the blue polygon outlines the node locations.

The performance of HPP is compared with two other approaches to establish a performance baseline and highlight the advantages and disadvantages of each approach. As previously mentioned, Kool uses a graph attention model (AM) to encode a learned representation of node locations with an attention layer to identify the most important nodes. With this representation, reinforcement learning is used to train an agent to identify the most optimal routes.

AM was trained for 3200 steps with a batch size of 256 for 100 epochs, which took approximately 8 hours on an NVIDIA 3080. The training became unstable for graph sizes larger than 100, and learning did not occur, so the trained models for 50 and 100 node graphs were used and compared.

The other method to be compared against HPP is Google's open-source linear optimization solver GLOP [8]. This method uses a primal-dual simplex algorithm to minimize the maximum route length give a distance matrix of ($n \times n$) nodes. Because of this exponential blow-up in complexity with the graph size, solutions for more than 200 nodes could not be found in a reasonable amount of time.

The dataset contains 500 graphs representing the locations of ponds in an aquaculture farm, divided into 100 instances of 100, 200, 300, 500, and 700 node graphs. An instance is built by laying a grid of points with equal spacing in an area of a randomly generated convex polygon and randomly deleting 20% of them for variation. The resulting graph's grid-like structure is expected to require different heuristics to successfully navigate than randomly placed nodes. One of the primary challenges for this application is the computational complexity required for solving VRP on large graphs. Few studies have analyzed graph data for more than 100 nodes, while this data contains up to 700 nodes. All methods attempt to divide the nodes between 5 routes. In practice, the number of drones to be used will be minimized depending on their maximum range.

We evaluate our algorithm on 3 metrics: average total distance, average maximum route distance, and run time to solve 100 instances. The average total distance will measure the method's overall efficiency, while the average maximum route distance tests each method's ability to distribute nodes evenly. Run time is difficult to fairly measure because of the differences in implementation, optimization and hardware. All experiments were run in Ubuntu 21.04 with an AMD

5600x CPU, NVIDIA RTX 3080 GPU, and 16 GB of 2133 MHz DDR4 RAM. HPP and GLOP utilize only the CPU, while AM's CUDA implementation can take advantage of parallel GPU structure. HPP[1] is implemented in MATLAB with no multi-threading, GLOP is primarily C++ and python, and AM is entirely python. Optimal VRP solutions do exist, but their long run times make them intractable for this application, so we cannot use the optimality gap as a performance metric.

It may be argued that for stationary locations, such as the ponds in an aquaculture farm, that the runtime is not important because the pathing will never change once an optimal solution is found. This is true, but the reality of operating aerial drones in windy conditions and changes in pond conditions (i.e., certain ponds in stress may need to be sampled at a higher frequency) may require the routing to be readjusted continuously to ensure successful operation. Should a drone be disabled or not reach a certain pond, the pathing algorithm must be able to run in real-time to correct any deficiencies. It is also vital to consider scaling the algorithm for commercial use, where a central server may simultaneously plan routes for several farms.

## 4. RESULTS

Table 1 shows the performance of HPP compared to AM trained on 50 and 100 node instances and GLOP. For graphs larger than 200 nodes, HPP is the best in terms of total distance, maximum route length and run time. However, GLOP's performance for graphs larger than 200 nodes was not considered because its run time was longer than 8 hours. The run time for HPP is orders of magnitude smaller than AM or GLOP, which is to be expected for a simple heuristic method. At only 100 nodes, GLOP takes nearly 11 minutes to solve 100 instances. This alone would disqualify GLOP as a practical solution to the HAUCS VRP. Interestingly, AM50 seems to outperform AM100 up to 100 nodes in terms of total distance and up to 200 nodes in terms of maximum route length, despite being trained on only 50 nodes. AM100 does outperform AM50 on larger graphs, especially on 700 node graphs where AM50 becomes unstable.

Table 1: The performance comparison of the proposed HPP against AM and GLOP

| **Average Total Distance** | | | | |
|---|---|---|---|---|
| Ponds | HPP | AM50 | AM100 | GLOP |
| 50 | 7.62 | 7.19 | 7.60 | **6.37** |
| 100 | 9.94 | 9.45 | 9.85 | **7.72** |
| 200 | 12.66 | 13.06 | 12.95 | **9.12** |
| 300 | **15.37** | 17.22 | 16.47 | --- |
| 500 | **19.02** | 25.46 | 22.79 | --- |
| 700 | **21.72** | 35.35 | 29.74 | --- |

| **Average Maximum Route Length** | | | | |
|---|---|---|---|---|
| Ponds | HPP | AM50 | AM100 | GLOP |
| 50 | 2.01 | 1.75 | 1.82 | **1.47** |
| 100 | 2.53 | 2.27 | 2.39 | **1.65** |
| 200 | 3.14 | 2.73 | 3.27 | **1.90** |
| 300 | **3.79** | 4.02 | 3.89 | --- |
| 500 | **4.58** | 5.57 | 6.32 | --- |
| 700 | **5.28** | 12.77 | 7.66 | --- |

---

[1] Source code for HPP is available at https://github.com/tonydavis629/HAUCS

| Run time for 100 instances (s) | | | | |
| --- | --- | --- | --- | --- |
| Ponds | HPP | AM50 | AM100 | GLOP |
| 50 | **1.38E-03** | 0.11 | 0.12 | 73.50 |
| 100 | **2.99E-03** | 0.35 | 0.31 | 658.15 |
| 200 | **4.76E-03** | 0.90 | 1.07 | 5435.26 |
| 300 | **9.51E-03** | 2.15 | 2.39 | --- |
| 500 | **1.07E-02** | 5.97 | 6.49 | --- |
| 700 | **1.55E-02** | 17.97 | 18.51 | --- |

Figure 3 illustrates the different style that HPP takes when producing a route solution. HPP uses the grid-like layout of the nodes by moving in straight lines wherever possible. Based on inspection of AM routes, it seems that AM did not learn this heuristic despite being trained on the HAUCS dataset. Pretrained models which used randomly placed nodes for training had a slightly worse performance than AM50 and AM100. It is certainly possible that a learning algorithm could learn such a heuristic and outperform HPP. GLOP's strategy appears to start all routes near the depot then circle around the graph to return to the depot at the end of the route, resulting in significantly shorter routes than either HPP or AM.

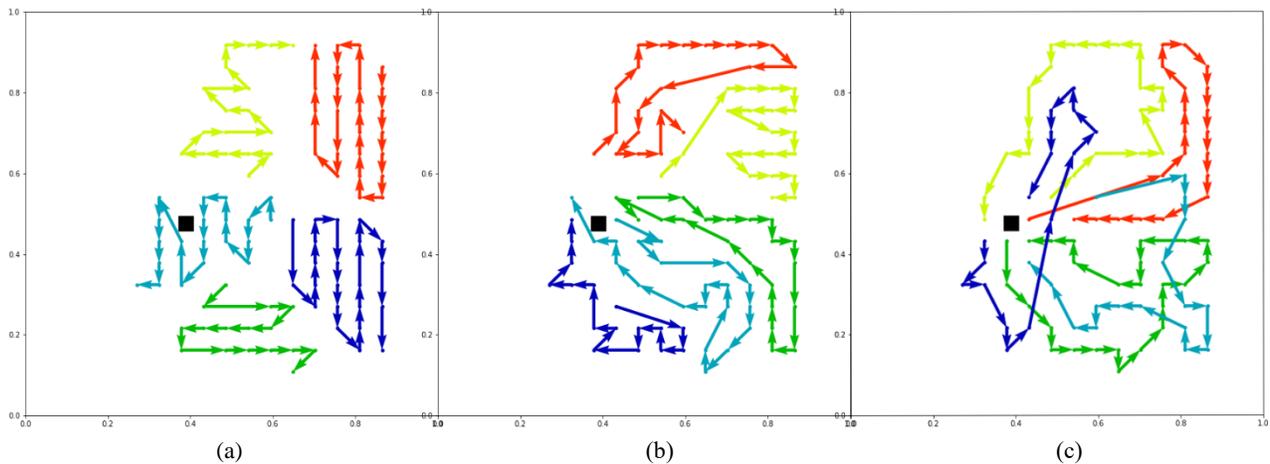

Figure 3: Route comparisons between (a) HPP, (b) AM and (c) GLOP. Depot is signified by the black box. Plotted using [22]

An additional benefit to HPP for aerial drones is the natural airspace deconfliction that results from assigning routes based on area, as seen in figure 3. While AM and GLOP may produce shorter routes in small graphs, they tend to cross each other's airspace, potentially resulting in air collisions. HPP routes are also more maintainable and interpretable for human operators. Should a problem occur during operation, it is easier to identify which drone failed based on its location.

## 5.  DISCUSSION

The ability for machine learning algorithms to develop heuristics for navigating routing problems has been proven to be successful. However, in most, if not all, investigations into machine learning algorithms' routing problem performance, the node locations have been distributed in a random fashion. In reality, problems typically have some kind of structure that can be exploited to develop simple and effective solutions whose performance are comparable to the most sophisticated learning algorithms. In this case, the grid-like distribution of nodes allows HPP to have comparable performance to AM because HPP is designed to move in straight lines down the rows and columns when possible, instead of diagonally. For unique settings such as our simulated aquaculture farm, hand-crafted heuristics like HPP may be more appropriate than learning algorithms because explicitly implementing human knowledge may be a more practical solution than what can be learned in training given current progress in machine learning. For complex environments where human intuition could not compete with learned heuristics, recent progress has shown that machine learning is a capable solution.

Additionally, it is important to remember that AM performs well in various routing problems and is not explicitly designed for this problem like HPP. So, while machine learning algorithms may not always be the most optimal, they may be the most generalizable solutions. This implies that machine learning algorithms are useful for navigating complex, unstructured data and can be used in various environments, but purpose-built classical algorithms are simpler to implement and can perform better depending on expert knowledge and the specific environment. However, the importance of flexible solutions cannot be understated, and it is plausible that machine learning algorithms could surpass all human written heuristics in the future.

Table 2: Summary of algorithm characteristics

|  | Optimality | Scalability | Practicality |
| --- | --- | --- | --- |
| HPP | Medium | High | High |
| AM | Medium | Medium | Medium |
| GLOP | High | Low | Medium |

On the other hand, numerical solutions such as GLOP can produce highly optimal routes at the expense of long run times. For all graph sizes that GLOP was able to compute, it produced the most optimal paths. As long as the run times do not interfere with reliable operations, this makes GLOP the more attractive solution for solving routes smaller than 200 nodes. However, airspace will have to more closely managed to prevent in-air collisions between the drones.

For larger graphs, HPP was the most practical solution because of its scalability. It achieved more optimal routes than AM in large graphs despite both solving in linear time. The benefit of the cluster-then-route heuristic only becomes apparent when enough nodes are included in a cluster. With too few nodes for each cluster, the back-and-forth routing cannot occur and the route is suboptimal. Table 2 summarizes the advantages and disadvantages of each method discussed.

Metaheuristic algorithms that focus on large graphs were also investigated, such as large neighborhood search, but their run times were significantly longer than GLOP, making them unsuited for our application. Additional research is necessary to thoroughly investigate the wide variety of VRP solutions available, especially those meant to solve large graphs.

## 6. CONCLUSION

One of the core components in the HAUCS framework is a team of cooperative robotic sensing platforms to support automated DO measurement. In this regard, an effective path planning algorithm to coordinate their operations is essential. We have presented the HAUCS Path Planning algorithm to plan routes for a team of aerial drones to measure dissolved oxygen levels in an aquaculture farm. The grid-like layout of an aquaculture farm allows a simple cluster-then-route heuristic to outperform some of the most sophisticated VRP solvers available on large graphs. Linear optimization solver GLOP produced the most optimal routes on smaller graphs. Modern operations research has focused on using metaheuristics and machine learning to solve a wide variety of problems, but it may be at the expense of creating simpler specialized solutions that are faster and more optimal for their specific application. HPP is also more amenable to hardware implementation because of its clear separation of airspace. The analysis of the various algorithms targeting different number of nodes is critical as well. There are various sizes of pond aquaculture farms, ranging from 100 or so ponds to 1000s of ponds. Therefore, in addition the scalability of a single algorithm for various farm sizes, it may also be important to consider the scalability of *a family of path planning algorithms*.

## 7. FUTURE WORK

The HAUCS project continues to develop the hardware and software integration necessary to realize the goal of automating DO monitoring in commercial aquaculture farms. Additional research is currently underway to establish the communication network, sensors, and charging stations required to make HAUCS possible. Additionally, it will be important to accurately predict DO levels several hours in the future to know the required measuring frequency for the drones.

As it is currently implemented, HPP is a method for the standard VRP, but due to the range limitations of drones and varying fish pond conditions throughout the day and seasons, the practical application of HAUCS will require a solution to the Distance Constrained VRP with Time Windows. To the best of our knowledge, this specific variation of the VRP has not been implemented and should be the subject of future research.

The position of the cluster centroids is essential to the optimality of HPP and is currently solved with k-means. If these clusters could be optimized such that each cluster's route is minimized, it would improve HPP's efficiency. Choosing optimal clusters is a combinatorial optimization problem in itself that may benefit from an additional layer of computation. Another potential improvement for HPP is to compare the back-and-forth routing against a more optimal heuristic or exact TSP solution.

## ACKNOWLEDGEMENT

This work was supported in part by the USDA-NIFA Grant 2019-67022-29204 and Aquaculture Specialty License Plate funds granted through the Harbor Branch Oceanographic Institute Foundation.